\documentclass[11pt,a4paper]{article}
\usepackage[colorlinks=true,linkcolor=blue]{hyperref}
\usepackage{graphicx}

\usepackage{cmap}
\usepackage[utf8x]{inputenc}
\usepackage[T1]{fontenc}
\usepackage[english]{babel}
\usepackage{amssymb,amsmath}
\usepackage{amsthm}
\usepackage{bbm}
\usepackage{microtype}
\usepackage{lmodern}

{
\rmfamily
\DeclareFontShape{T1}{lmr}{m}{sc}{<->ssub*cmr/m/sc}{}
\DeclareFontShape{T1}{lmr}{b}{sc}{<->ssub*cmr/b/sc}{}
\DeclareFontShape{T1}{lmr}{bx}{sc}{<->ssub*cmr/bx/sc}{}
}

\newcommand{\thmheadercommand}[1]{\textbf{\scshape{}#1.\\*}}

\newtheoremstyle{yannthm}{\topsep}{\topsep}{\slshape}{}{\scshape\bfseries}{.}{.5em}{%
\thmname{#1}\thmnumber{ #2}\thmnote{#3}%
}
\newtheoremstyle{yannthm2}{\topsep}{\topsep}{}{}{\scshape\bfseries}{.}{.5em}{%
\thmname{#1}\thmnumber{ #2}\thmnote{#3}%
}

\def\d{\operatorname{d}\!{}}

\def\R{{\mathbb{R}}}

\renewcommand{\geq}{\geqslant}
\renewcommand{\leq}{\leqslant}

\renewcommand{\emptyset}{\varnothing}

\newcommand{\deq}{\mathrel{\mathop:}=}

\def\eps{\varepsilon}
\renewcommand{\epsilon}{\varepsilon}
\renewcommand{\phi}{\varphi}

\let\oldPr\Pr
\renewcommand{\Pr}{\oldPr\nolimits}
\newcommand{\E}{\mathbb{E}}

\DeclareMathOperator{\Id}{Id}

\DeclareMathOperator{\diag}{diag}

\newcommand{\abs}[1]{\left\lvert#1\right\rvert}
\newcommand{\norm}[1]{\left\lVert#1\right\rVert}

\newenvironment{dem}[1][]{\begin{proof}[\thmheadercommand{Proof#1}]~\newline\ignorespaces}{\end{proof}}

{
\theoremstyle{yannthm}
\newtheorem{defi}{Definition}
\newtheorem*{defi*}{Definition}
\newtheorem{prop}[defi]{Proposition}
\newtheorem*{prop*}{Proposition}
\newtheorem{thm}[defi]{Theorem}
\newtheorem*{thm*}{Theorem}
\newtheorem{lem}[defi]{Lemma}
\newtheorem*{lem*}{Lemma}
\newtheorem{cor}[defi]{Corollary}
\newtheorem*{cor*}{Corollary}

\newtheorem*{ex*}{Example}

\newtheorem*{subenonce}{}

\theoremstyle{yannthm2}

\newtheorem*{exo*}{Exercise}

\newtheorem*{rem*}{Remark}

\newtheorem*{subenonce2}{}

}

\newcommand{\transp}[1]{#1^{\!\top}\!}

\title{True Asymptotic Natural Gradient Optimization}

\author{Yann Ollivier}
\date{}

\newcommand{\D}{\mathcal{D}}
\newcommand{\deltat}{\operatorname{\delta}\!\hspace{-.03em}t}
\renewcommand{\transp}[1]{#1^{\scriptscriptstyle\top}}
\newcommand{\opnorm}[1]{\norm{#1}_{\mathrm{op}}}

\begin{document}

\maketitle

\begin{abstract}
We introduce a simple algorithm, True Asymptotic Natural Gradient
Optimization (TANGO), that converges to a true natural gradient
descent in the limit of small learning rates, without explicit Fisher matrix
estimation.

For quadratic models the algorithm is also an instance of averaged
stochastic gradient,
where the parameter is a moving average of a ``fast'',
constant-rate gradient descent. TANGO appears as a particular
de-linearization of averaged SGD, and is sometimes quite
different on non-quadratic models. This further connects 
averaged SGD and natural gradient, both of which are arguably
optimal asymptotically.

In large dimension, small learning rates will be required to approximate
the natural gradient well. %, so the practical interest for natural gradient
%is not clear. 
Still, this shows it is possible to get arbitrarily close to exact
natural gradient descent with a lightweight algorithm.
\end{abstract}

Let $p_\theta(y|x)$ be a probabilistic model for predicting output values $y$
from inputs $x$ ($x=\emptyset$ for unsupervised learning). Consider the associated log-loss
\begin{equation}
\ell(y|x)\deq -\ln p_\theta(y|x)
\end{equation}
Given a dataset 
$\D$ of pairs $(x,y)$, we optimize the average log-loss
over $\theta$
via a momentum-like gradient descent.

\begin{defi}[ (TANGO)]
Let $\deltat_k\leq 1$ be a sequence of learning rates and let $\gamma>0$. Set
$v_0=0$. Iterate the following: 
\begin{itemize}
\item Select a sample $(x_k,y_k)$ at random in
the dataset $\D$.

\item Generate a pseudo-sample $\tilde y_k$ for input $x_k$ according
to the predictions of the current model, $\tilde y_k\sim
p_\theta(\tilde y_k|x_k)$ (or just
$\tilde y_k=y_k$ for the ``outer product'' variant). Compute gradients
\begin{equation}
g_k\gets \frac{\partial \ell(y_k|x_k)}{\partial \theta},\qquad \tilde g_k \gets
\frac{\partial \ell(\tilde y_k|x_k)}{\partial \theta}
\end{equation}

\item Update the velocity and parameter via
\begin{align}
v_k&= (1-\deltat_{k-1})v_{k-1}+\gamma g_k - \gamma (1-\deltat_{k-1})(\transp{v_{k-1}}\,\tilde g_k)\tilde g_k
\label{eq:v}
\\
\theta_k&= \theta_{k-1} - \deltat_k v_k \label{eq:thetaupdate}
\end{align}
\end{itemize}
\end{defi}

TANGO is built to approximate Amari's \emph{natural gradient} descent,
namely, a gradient descent preconditioned by the inverse of the Fisher
information matrix of the probabilistic model $p_\theta$ (see definitions
below).  The natural gradient arguably provides asymptotically optimal
estimates of the parameter $\theta$
\cite{Amari1998}. However, its use is unrealistic for large-dimensional models
due to the computational cost of storing and inverting the Fisher matrix,
hence the need for approximations.
One of its key features is its invariance to any change of variable in
the parameter $\theta$ (contrary to simple gradient descent). The natural gradient
is also a special case of the \emph{extended Kalman filter} from estimation
theory \cite{natkal}, under mild conditions.

In TANGO, $\deltat/\gamma$ should be small for a good natural gradient
approximation.

For stability of the update \eqref{eq:v} of $v$, $\gamma$ should be taken small
enough; but a small $\gamma$ brings slower convergence to the natural
gradient. A conservative, theoretically safe choice is setting $\gamma=1/\max
\norm{\tilde g}^2$ using the largest norm of $\tilde g$ seen so far. This
may produce a too small $\gamma$ if gradients are unbounded.
If the gradients follow a Gaussian distribution (with any covariance
matrix), then
$\gamma=1/\E[3\norm{\tilde g}^2]$ is theoretically safe; the average can
be estimated on past gradients. In general, $\gamma\leq \E[\norm{\tilde
g}^2]/\E[\norm{\tilde g}^4]$ is a necessary but not sufficient condition;
this may be used as a starting point. (See discussion after
Theorem~\ref{thm:general}.)

\begin{figure}
\centerline{\includegraphics[width=.6\textwidth]{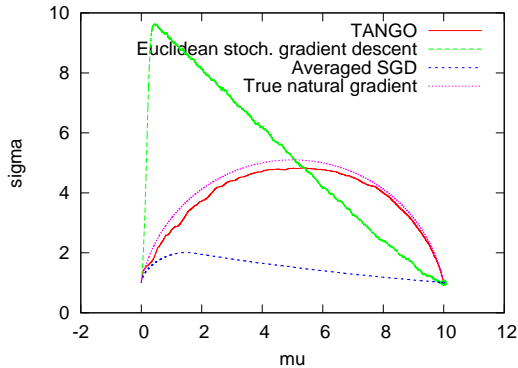}}
\caption{Learning a Gaussian model $\mathcal{N}(\mu,\sigma^2)$ with
unknown $\mu$ and $\sigma$, via gradient descent on $(\mu,\ln\sigma)$.
The initial point is $\mathcal{N}(0,1)$ and the data are
$\mathcal{N}(10,1)$. The Fisher metric is isometric to the hyperbolic
plane $(\mu,\sigma)$, whose geodesics are circles, so that the true natural
gradient starts by increasing variance so that $\mu$ moves faster.
Plotted are trajectories of SGD with learning rate $10^{-3}$, and TANGO
and averaged SGD with $\gamma=10^{-2}$ and $\deltat=10^{-4}$.}
\label{fig:gaussian}
\end{figure}

TANGO enjoys the following properties:
\begin{enumerate}
\item TANGO converges to an \emph{exact} natural gradient trajectory when
the learning rate $\deltat$ tends to $0$ with $\gamma$ fixed, namely, to the trajectory of the ordinary
differential equation $\d\theta/\d t=-J(\theta)^{-1} \E[\partial
\ell/\partial\theta]$ with $J$ the Fisher matrix at $\theta$ (Theorem~\ref{thm:tango}).
\item For $\deltat=1$ TANGO is an ordinary gradient descent with constant
learning rate $\gamma$.
% \item
% When both $\deltat$ and $\gamma$ tend to $0$ with $\deltat/\gamma$ fixed,
% TANGO converges to trajectories of a regularized natural gradient descent
% $\d\theta/\d t=-(J(\theta)+\frac{\deltat}{\gamma}\Id)^{-1} \E[\partial
% \ell/\partial\theta]$. In particular, $\deltat$ must be smaller than
% $\gamma$ to reach a correct natural gradient approximation.
\item For quadratic losses, TANGO is an instance of \emph{averaged
stochastic gradient descent}
with additional noise (Proposition~\ref{prop:trajav}): a ``fast'' stochastic gradient descent
with constant learning rate is performed, and the algorithm returns a
moving average of this trajectory (updated by a factor $\deltat_k$ at
each step). However, for non-quadratic losses,
TANGO can greatly differ from averaged SGD
(Fig.~\ref{fig:gaussian}).
\end{enumerate}

Thus, TANGO smoothly interpolates
between ordinary and natural gradient descent when the learning rate
decreases.

To illustrate the convergence to the natural gradient in an informal way,
take $\deltat=0$. Then $\theta$ does not move, and the average of $g$ is
the gradient of the expected loss at $\theta$. Then the average
of $v$ over time
converges to $(\E \tilde g\transp{\tilde g})^{-1}\E g$, the exact natural
gradient direction at $\theta$. Indeed, this is the only fixed point of
\eqref{eq:v} in expectation. Actually, \eqref{eq:v} is a way of solving
for $(\E \tilde g\transp{\tilde g})v=\E g$ by stochastic gradient
descent on $v$. The Fisher matrix $J$ is $\E \tilde g\transp{\tilde g}$ by
definition.

\paragraph{Acknowledgments.} The author would like to thank Léon Bottou,
Guillaume Charpiat, Fabrice Debbasch, Aaron Defazio, Gaétan Marceau-Caron
and Corentin Tallec for helpful discussions and comments around these ideas.

\paragraph{Related work.}
Three different lines of work lead to TANGO-like algorithms. Averaged SGD
\cite{polyakjuditsky1992,ruppert1988} uses a ``fast'' gradient descent
with large learning rate
$\gamma$ (here on the variable $v$), with an averaging
operation on top (here by accumulation into $\theta$).
For linear problems $\gamma$ can be kept
constant.

Averaged SGD achieves the asymptotically optimal Cramer--Rao
bound involving the inverse Fisher matrix, although ``no explicit Hessian inversion has been performed''
\cite{bachmoulines2011,polyakjuditsky1992}. TANGO may clarify how the
implicit Hessian or Fisher matrix inversion occurs.

Later work on averaged SGD focussed on non-asymptotic behavior
(especially, forgetting of the starting point),
on somewhat dimension-independent
bounds, and on larger $\gamma$ for linear models \cite{bachmoulines2011,
bachmoulines2013, martens2014natgradinsights,defossezbach2015, dieuleveut2016}.  A constant, large
$\gamma$ provides the most benefits; yet for nonlinear models, averaged
SGD with constant
$\gamma$ leads to biases, hence the need for TANGO.
Our analysis of the dynamics of $v$ in TANGO and in
Theorem~\ref{thm:general} below follows this line of work.

Previous work on approximating the natural gradient for
large-dimensional models, such as TONGA and others
\cite{TONGA,Ollivier2015,martensgrosse2015,desjardins2015NNN,marceauollivier2016}, 
did not provide an arbitrarily good approximation to the Fisher matrix,
as it relied
on structural matrix approximations
(diagonal, block-diagonal, diagonal plus small-rank...)
An
exception is \cite{desjardins2013metricfreeNG} for Boltzmann machines,
directly transposed from the Hessian-free Newton method of
\cite{Martens2010,Martens2011,Martens2012}: at each step, a
large number of auxiliary conjugate gradient steps are performed to solve for Fisher
matrix inversion, before the main update of the parameter occurs. From
this viewpoint, TANGO performs the main gradient descent on $\theta$ and
the auxiliary gradient descent at the same time.

For quasi-Newton methods in the convex case, auxiliary gradient descents
to approximate the inverse Hessian have been suggested several times; see
\cite{agarwal2016secondorderlineartime,Martens2010,Martens2011,Martens2012}
and the references therein.  Second-order methods for neural networks
have a long history, see e.g.\ \cite{lecun98efficientbackprop}.
\footnote{Technically the natural gradient is not a second-order method,
as the Fisher matrix represents a Riemannian metric tensor rather than a Hessian of the
loss. It can be
computed from squared gradients, and the natural gradient is well-defined
even if the loss is flat or concave. The Fisher matrix coincides with the
Hessian of the loss function only asymptotically at a local minimum,
provided the data follow the model.}

Third, ``two-timescale'' algorithms in reinforcement learning use updates
reminiscent of TANGO, where the ``fast'' timescale is used to approximate
a value function over a linear basis via a least squares method, and the
``slow'' timescale is used to adapt the parameters of a policy. For
instance, the main results of \cite{tadic2004twoscales} or
\cite{karmakar2017twotimescaletd} deal with convergence of updates
generalizing
\eqref{eq:v}--\eqref{eq:thetaupdate}.  However, these results
crucially assume that both $\deltat$ and $\gamma$ tend to $0$. This would
be too slow in our setting. A constant $\gamma$ can be used in TANGO
(and in averaged SGD for linear least squares) thanks to the linearity of the
update of $v$, but this requires a finer analysis of noise.

\paragraph{Discussion and shortcomings.}
Critical to TANGO is the choice of the parameter $\gamma$: the larger
$\gamma$ is, the faster the trajectory will resemble natural gradient (as
$v$ converges faster to $(\E \tilde g\transp{\tilde g})^{-1} \E g$).
However, if $\gamma$ is too large the update for $v$ is numerically
unstable. For averaged SGD on quadratic losses, the choice of $\gamma$
is theoretically well understood \cite{defossezbach2015}, but the situation is
less clear for non-quadratic losses. We provide some general guidelines
below.

The algorithmic interest of
using TANGO with respect to direct Fisher matrix computation is not
clear. Indeed, for $\deltat=0$, the update equation \eqref{eq:v} on $v$
actually solves
$v=(\E \tilde g\transp{\tilde g})^{-1} \E g$ by stochastic gradient
descent on $v$. 
The speed of convergence is heavily dimension-dependent,
a priori. Similar Hessian-free Newton algorithms that rely on an auxiliary
gradient descent to invert the Hessian, e.g., \cite{Martens2010}, need a large number of auxiliary
gradient iterations.
In this case, the interest of TANGO may be its ease of implementation.

Still, averaged SGD is proved to accelerate convergence for
quadratic problems \cite{polyakjuditsky1992}. So TANGO-like algorithms bring
benefits in some regimes. 

For linear models, \cite{dieuleveut2016} study situations in
which the convergence of \eqref{eq:v} happens faster than suggested by
the dimension of the problem, depending on the eigenvalues of the
Hessian. For non-linear problems, this may be the case if the data
clusters naturally in a few groups (e.g., classification with few
labels): sampling a value of $\tilde y$ in each of the clusters may
already provide an interesting low-rank approximation of the Fisher
matrix $\E \tilde g\transp{\tilde g}$. In such a situation, $v$ may
converge reasonably fast to an approximate natural gradient direction.

\paragraph{Implementation remarks: minibatches, preconditioned TANGO.}
If $\tilde g$ is computed as the average over a minibatch of size $B$,
namely
$\tilde g=\frac{1}{B} \sum_{i=1}^B \tilde g_i$ with $\tilde g_i$ the
gradient corresponding to output sample $\tilde y_i$ in the minibatch,
then the equation for $v$ has to be modified to
\begin{equation}
v_k= (1-\deltat_{k-1})v_{k-1}+\gamma g_k - \gamma B
(1-\deltat_{k-1})(\transp{v_{k-1}}\,\tilde g_k) \tilde g_k
\end{equation}
because the expectation of $\tilde g\transp{\tilde g}$ is $\frac{1}{B}$
times the Fisher matrix.

Preconditioned TANGO (e.g., à la RMSProp) can be obtained by choosing a
positive definite matrix $C$ and iterating
% \footnote{
% Or equivalently
% \begin{align}
% v_k&= (1-\deltat_{k-1})v_{k-1}+\gamma g_k - \gamma
% (1-\deltat_{k-1})(\transp{v_{k-1}}\,C \tilde g_k) \tilde g_k
% \\
% \theta_k&= \theta_{k-1} - \deltat_k C v_k
% \end{align}
% which is equivalent but suggests to learn $C$ from the $v$'s rather than
% the $g$'s. TODO: check in practice... Probably a very bad idea since $Cv$
% converges to the nat grad direction, so adapting $C$ to try to make it
% converge to something else will be a catastrophe.
% }
\begin{align}
v_k&= (1-\deltat_{k-1})v_{k-1}+\gamma C g_k - \gamma
(1-\deltat_{k-1})(\transp{v_{k-1}}\,\tilde g_k) C\tilde g_k
\\
\theta_k&= \theta_{k-1} - \deltat_k v_k
\end{align}
(This is TANGO on the variable $C^{-1/2}\theta$.)
The matrix $C$ may help to improve conditioning of gradients and of the
matrix $C\E\tilde g\transp{\tilde g}$. Choices of $C$ may include RMSProp
(the
entrywise reciprocal of the root-mean-square average of gradients) or the inverse of
the diagonal Fisher matrix, $C^{-1}=\diag (\E \tilde g^{\odot 2})$. These
options
will require different adjustements for $\gamma$.

Quadratic output losses can be seen as the log-loss of a probabilistic
model, $\ell(y|x)=\frac{\norm{y-f_\theta(x)}^2}{2\sigma^2}$ for any
value of $\sigma^2$. However, $\sigma^2$ should be set to the actual mean
square error on the outputs, for the natural gradient descent to work
best. The choice of $\sigma^2$ affects both the scaling of gradients $g$ and $\tilde g$, and
the sampling of pseudo-samples $\tilde y$, whose law is
$\mathcal{N}(f_\theta(x),\sigma^2)$.

\paragraph{TANGO as an instance of averaged SGD for quadratic losses.}
Averaged SGD maintains a fast-moving parameter with constant
learning rate, and returns a moving average of the fast trajectory. It is
known to have excellent asymptotic properties for quadratic models.

For quadratic losses, TANGO can be rewritten as a form of averaged SGD,
despite TANGO only using gradients evaluated at the ``slow''
parameter $\theta$. This is specific to gradients being a linear function
of $\theta$.

Thus TANGO can be considered as a non-linearization of averaged SGD,
written using gradients at $\theta$ only. Even for
simple nonlinear models, the difference can be substantial
(Fig.~\ref{fig:gaussian}). For nonlinear models, averaged SGD
with a fixed learning rate $\gamma$ can have a bias of size comparable to
$\gamma$, even with small $\deltat$.
%For instance, on
%Fig.~\ref{fig:gaussian}, trajectory averaging systematically converges to
%a model with variance $\approx 1.1$ instead of $1$.
\footnote{A bias of
size $\gamma$ is easy to see on the following example: Define a loss
$\ell(x)=\abs{x}$ for $\abs{x}\geq \gamma/2$, and extend this loss in an
arbitrary way on the interval $[-\gamma/2;\gamma/2]$. Since the gradients
are $\pm 1$ out of this interval, a gradient descent with fixed learning
rate $\gamma$, initialized at a multiple of $\gamma/2$, will make jumps
of size exactly $\gamma$ and never visit the interior of the interval
$[-\gamma/2;\gamma/2]$. Whatever the average parameter of this
trajectory is, it is unrelated to the behavior of the loss on
$[-\gamma/2;\gamma/2]$ and to the location of the minimum. Thus
averaged SGD can have a bias of size $\approx \gamma$, whatever
$\deltat$.} TANGO does not exhibit such a bias.

\newcommand{\thetaf}{\theta^{\mathrm{fast}}}
\newcommand{\gf}{g^{\mathrm{fast}}}

\begin{prop}
\label{prop:trajav}
Assume that for each sample $(x,y)$, the log-loss $\ell(y|x)$ is a quadratic
function of $\theta$ whose Hessian does not depend on $y$ (e.g., linear
regression $\ell(y|x)=\frac12 \norm{y-\smash{\transp{\theta}}x}^2$).

Then TANGO is identical to the following trajectory
averaging algorithm:
\begin{align}
%\thetaf_k&=\thetaf_{k-1}-\gamma\left.\frac{\partial \ell(y_k|x_k)}{\partial \theta}\right|_{\theta=\thetaf_{k-1}}+\xi_k
\thetaf_k&=\thetaf_{k-1}-\gamma\frac{\partial \ell(y_k|x_k)}{\partial
\thetaf_{k-1}}+\gamma \xi_k
\\
\theta_k &= (1-\deltat_k)\theta_{k-1}+\deltat_k \thetaf_k
\end{align}
where $\xi_k$ is some centered random variable whose law depends on
$\thetaf_{k-1}$ and $\theta_{k-1}$.
The identification with TANGO is via $v_k=\theta_{k-1}-\thetaf_k$.
\end{prop}

The proof (Appendix~\ref{sec:trajavproof}) is mostly by direct algebraic
manipulations. For quadratic losses, the gradients are a linear function
of the parameter, so that the derivative at point $\thetaf$ can be
rewritten as the derivative at point $\theta$ plus a Hessian term; for
quadratic losses, the Hessian is equal to the Fisher metric.

The additional noise $\xi_k$ is multiplicative in $v$. This is standard for
linear regression \cite{dieuleveut2016}: indeed, in linear
regression, the gradient from sample $(x,y)$ is $-yx+x\transp{x}\theta$,
and its expectation is $-\E(yx)+\E(x\transp x)\theta$ so that the gradient
noise has a multiplicative component $(x\transp{x}-\E(x\transp
x))\theta$. (Treatments of gradient
descent often assume additive noise instead, see discussion in
\cite{dieuleveut2016}.)

Replacing the TANGO update of $\theta$ in
\eqref{eq:thetaupdate} with $\theta_k=\theta_{k-1}-v_k$ would make TANGO
equivalent to an \emph{accelerated gradient} method with additional noise for
quadratic functions.%  The additional noise
% stems from using a Hessian estimate $\tilde g_k\transp{\tilde g_k}$ instead of
% the exact Hessian $\E \tilde g_k\transp{\tilde g_k}$.

\paragraph{Convergence of TANGO to the natural gradient.} Let the Fisher matrix of the model be
\begin{equation}
J(\theta)\deq \E \tilde g\transp{\tilde g}=\E_{(x,y)\in \D} \E_{\tilde
y\sim p_\theta(\tilde y |x)}\frac{\partial \ell(\tilde y|x)}{\partial
\theta}^{\otimes 2}
\end{equation}
where, for a column vector $v$, $v^{\otimes 2}$ is the outer product
$v\transp{v}$.

% Basically, the update for $v$ solves for $v=J^{-1}g$ via a stochastic gradient
% descent on $g-Jv$ using $J=\E \tilde g \transp{\tilde g}$. I do not expect this process to converge fast... it is
% rather similar to the auxiliary conjugate gradient descent in Hessian-free...

The stochastic natural gradient descent on $\theta$, with learning rate
$\deltat$, using the exact Fisher matrix $J(\theta)$, is
\begin{equation}
\theta^{t+\deltat}=\theta^t-\deltat J(\theta^t)^{-1} \frac{\partial
\ell(y_k|x_k)}{\partial \theta^t}
\end{equation}
where at each step $(x_k,y_k)$ is a random sample from the
dataset $\D$. In the limit of small learning rates $\deltat\to 0$, it
converges to a ``true'' continuous-time natural gradient descent
trajectory, driven by the differential equation
\begin{equation}
\label{eq:truenat}
\frac{\d \theta^t}{\d t}=-J(\theta^t)^{-1} \E_{(x,y)\in D} \frac{\partial
\ell(y|x)}{\partial \theta^t}
\end{equation}

\begin{thm}
\label{thm:tango}
Make the following regularity assumptions: The second moment of gradients
$g$ is bounded over $\theta$. The fourth moment of gradients
$\tilde g$ is bounded over $\theta$. The lowest eigenvalue of
the Fisher matrix $J(\theta)$, as a function of $\theta$, is bounded away
from $0$. The Fisher matrix is a $C^1$ function of $\theta$ with bounded
first derivatives.

Let $\theta^T$ be the value of the exact natural gradient
\eqref{eq:truenat} at time $T$. Assume that the parameter $\gamma$ in
TANGO is smaller than some
constant that depends on the moments of the gradients and the eigenvalues of
the Fisher matrix.

Then
the value of $\theta$ obtained
after $T/\deltat$ iterations of TANGO converges in probability to $\theta^T$,
when
$\deltat\to 0$.% with $\deltat \ll \gamma$.
\end{thm}

The probability in this theorem refers to the random choice of samples
$x_k$, 
$y_k$ and $\tilde y_k$ in TANGO.

Theorem~\ref{thm:tango} will be obtained as a corollary of the more general
Theorem~\ref{thm:general}, which also provides quantitative versions of
the choice of $\gamma$ in TANGO.

To illustrate a key idea of the proof, we start with a simpler,
noise-free situation.

\newcommand{\lambdamin}{\lambda_\mathrm{min}}
\newcommand{\lambdamax}{\lambda_\mathrm{max}}

\begin{prop}
\label{prop:nonstoch}
Consider the iteration of
\begin{align}
v_k&= v_{k-1}+\gamma F(\theta_{k-1}) - \gamma A(\theta_{k-1})v_{k-1}
\\
\theta_k&= \theta_{k-1} - \deltat \,v_k
\end{align}
initialized at $v_0=0$, where $F$ is a vector field on $\theta$ and $A$ is a 
field of symmetric positive definite matrices.

Assume that $F$ and $A$ are $C^1$ with bounded derivatives. Let
$\lambdamin\deq \inf_\theta \min \mathrm{eigenvalues}(A(\theta))$ and 
$\lambdamax\deq \sup_\theta \max \mathrm{eigenvalues}(A(\theta))$, and assume
$\lambdamin>0$ and $\lambdamax<\infty$.
Fix $\gamma$ smaller than
$1/\lambdamax$.

Then
when $\deltat\to 0$, the value $\theta$ of this system after $T/\deltat$
iterations converges to the value at time $T$ of the
ordinary differential equation with preconditioning $A^{-1}$,
\begin{equation}
\frac{\d \theta^t}{\d t}=-A(\theta^t)^{-1} F(\theta^t)
\end{equation}
initialized at $\theta^0=\theta_0$. More precisely,
$\theta_{T/\deltat}-\theta^T=O(\deltat)$.
\end{prop}

\begin{dem}
We first deal with the case of constant $A(\theta)\equiv A$.

First, note that the sums of the contributions of $v_1$ to all future
updates of $\theta$ is $\deltat \sum (\Id-\gamma A)^kv_1=\deltat
\gamma^{-1}A^{-1}v_1$.

This suggests setting
\begin{equation}
z_{k+1}\deq \theta_k-\deltat \gamma^{-1}A^{-1}v_{k+1}
\end{equation}
which contains ``$\theta_k$ plus all the known future updates from the
terms $F(\theta_j)$, $j\leq k$, that are  already present in $v_k$''. 
Substituting for $\theta_k$ and $v_{k+1}$ in $z_{k+1}$, one finds that
the update for $z$ is
\begin{align}
z_{k+1}&= \theta_{k-1}-\deltat v_k-\deltat\gamma^{-1}A^{-1}(
v_k+\gamma F(\theta_k)-\gamma A v_k)
\\&=z_k-\deltat A^{-1}F(\theta_k)
\end{align}
which only involves the new contribution from $F(\theta_n)$, and not
$v$.

Moreover, 
\begin{equation}
z_k=\theta_{k-1}-\deltat\gamma^{-1}A^{-1}v_k=
\theta_k+\deltat\, v_k -\deltat \gamma^{-1}A^{-1}v_k=\theta_k+O(\deltat
\norm{v_k})
\end{equation}
since $A^{-1}$ is bounded (its largest eigenvalue is $1/\lambdamin$).

Now, the update for $v_k$ is $(1-\gamma\lambdamin)$-contracting, because
the condition
$\gamma<1/\lambda_\mathrm{max}$ implies that the eigenvalues of
$\gamma A$ lie between $\gamma\lambdamin$ and $1$. Since $\lambdamin>0$ and $F$ is
bounded, it is easy to show by induction that $\norm{v_k}\leq (\sup
\norm{F})/\lambdamin$ so that $v$ is bounded.

Therefore, $z_k=\theta_k+O(\deltat)$.
Then, given the regularity assumptions on
$F$, one has
$F(\theta_k)=F(z_k)+O(\deltat)$ and
\begin{equation}
z_{k+1}=z_k-\deltat A^{-1}F(z_k)+
O(\deltat^2)
\end{equation}
since $A^{-1}$ is bounded.
This does not involve $v$ any more.

But this update for $z_k$ is just a Euler numerical scheme for the differential equation $\dot
z=- A^{-1}F(z)$. So by the standard theory of approximation of ordinary
differential equations, when $\deltat\to 0$, $z_{T/\deltat}$ converges
to the solution at time $T$ of this equation, within an error $O(\deltat)$. Since 
$\theta_k-z_k$ is $O(\deltat)$ as well, we get the same conclusion for $\theta$.

For the case of variable $A$, set
\begin{equation}
z_{k+1}\deq \theta_k-\deltat\gamma^{-1}A^{-1}(\theta_k)v_{k+1}
\end{equation}
and substituting for $\theta_k$ and $v_{k+1}$ in this definition, one
finds
\begin{align}
z_{k+1}&= \theta_{k-1}-\deltat \gamma^{-1} A(\theta_k)^{-1} v_k-\deltat
A(\theta_k)^{-1}F(\theta_k)
\\&=
z_k+\deltat\gamma^{-1}(A(\theta_{k-1})^{-1}-A(\theta_k)^{-1})v_k-\deltat
A(\theta_k)^{-1}F(\theta_k)
\end{align}

Now, under our eigenvalue assumptions, $A^{-1}$ is bounded. Since $A$ has bounded derivatives, so does
$A^{-1}$ thanks to $\partial_{\theta} A^{-1}=-A^{-1}(\partial_\theta
A)A^{-1}$. Therefore we can apply a Taylor expansion of $A^{-1}$ so that
\begin{equation}
A(\theta_{k-1})^{-1}-A(\theta_k)^{-1}=O(\theta_{k-1}-\theta_k)=O(\deltat\norm{v_k})
\end{equation}
so that
\begin{equation}
z_{k+1}=z_k-\deltat
A(\theta_k)^{-1}F(\theta_k)+O(\deltat^2\norm{v_k}^2)
\end{equation}
after which the proof proceeds as for the case of constant $A$, namely:
$z_k-\theta_k$ is $O(\deltat\norm{v_k})$ so that
\begin{equation}
z_{k+1}=z_k-\deltat
A(z_k)^{-1}F(z_k)+O(\deltat^2\norm{v_k}+\deltat^2\norm{v_k}^2)
\end{equation}
and $v_k$ is bounded by induction. So the update for $z_k$ is a Euler
numerical scheme for the differential equation $\dot z=-A(z)^{-1}F(z)$,
which ends the proof.
\end{dem}

% Sketch of proof: Let $g_k$ be the gradient using at iteration $k$. After
% $k$ iterations, the value of $v$ is
% \begin{equation}
% v=\sum_{i=1}^k (\Id-\gamma C)^{k-i} \gamma g_i
% \end{equation}
% and therefore the value of $\theta$ after $K$ iterations is
% \begin{align}
% \theta&=\theta_0+\deltat \sum_{k=1}^K \sum_{i=1}^k (\Id-\gamma C)^{k-i}
% \gamma g_i
% \\&=
% \theta_0+\deltat \sum_{i=1}^K \sum_{k=i}^K (\Id-\gamma C)^{k-i}
% \gamma g_i
% \\&= \theta_0+\deltat \sum_{i=1}^K \left(\sum_{k=0}^{K-i} (\Id-\gamma
% C)^k\right)
% \gamma g_i
% \end{align}
% 
% When $K\to\infty$, the sum $\sum (\Id-\gamma C)^k$ tends to
% $\left(\Id -(\Id-\gamma C)\right)^{-1}=\gamma^{-1}C^{-1}$. The condition
% on $\gamma$ compared to the eigenvalues of $C$ ensures convergence. Therefore when $K\to \infty$ each term $g_i$
% contributes $\deltat C^{-1}g_i$ to $\theta$. TODO: clean limit
% $\deltat\to 0$ and $K=T/\deltat$, and $g_i$ depends on the trajectory
% itself because it is a gradient at the current parameter value.

\newcommand{\filtr}{\mathcal{F}}

We now turn to the stochastic version of Proposition~\ref{prop:nonstoch}.
This provides a generalization of Theorem~\ref{thm:tango}:
Theorem~\ref{thm:tango} is a corollary of Theorem~\ref{thm:general} using
$\hat F_k=g_k$ and 
$\hat A_k=(1-\deltat) \tilde g_k\transp{\tilde
g_k}+\frac{\deltat}{\gamma} \Id$.

For numerical simulations of stochastic differential equations, the usual
rate of convergence is $O(\sqrt{\deltat})$ rather than $O(\deltat)$
\cite{KloedenPlaten92}. 

\begin{thm}
\label{thm:general}
Consider the iteration of
\begin{align}
v_k&= v_{k-1}+\gamma \hat F_k - \gamma \hat A_k v_{k-1}
\\
\theta_k&= \theta_{k-1} - \deltat \,v_k
\end{align}
initialized at $v_0=0$, where $\hat F_k$ is a vector-valued random variable
and $\hat A_k$ is a symmetric-matrix-valued random variable.

Let $\filtr_k$ be the sigma-algebra
generated by all variables up to time $k$, and abbreviate $\E_k$ for
$\E [\,\cdot \mid \filtr_k ]$. Let
\begin{equation}
F_k\deq \E_{k-1}\hat F_k,\qquad
%A_k\deq \E\left[\frac12 (\hat A_k+\transp{\hat A_k})\mid \filtr_{k-1}\right]
A_k\deq \E_{k-1}\hat A_k
\end{equation}
and assume that these depend on $\theta_{k-1}$ only, namely, that
exist functions $F(\theta)$ and $A(\theta)$ such that
\begin{equation}
F_k=F(\theta_{k-1}),\qquad A_k=A(\theta_{k-1})
\end{equation}
Assume that the functions $F$ and $A$ are $C^1$ with bounded derivatives.
Let $\lambda\deq \inf_\theta \min \mathrm{eigenvalues}(A(\theta))$, and assume $\lambda>0$.

Assume the following variance control:
for some $\sigma^2\geq 0$ and $R^2\geq 0$,
\begin{equation}
\E_{k-1}\norm{\hat F_k}^2\leq \sigma^2
,\qquad
\E_{k-1}\left[\transp{\hat A_k}\hat A_k \right]\preccurlyeq
R^2 A_k
\end{equation}
where $A\preccurlyeq B$ means $B-A$ is positive semidefinite.

Fix $0< \gamma\leq 1/R^2$.

Then
when $\deltat\to 0$, the value $\theta$ of this system after $T/\deltat$
iterations converges in probability to the value at time $T$ of the
ordinary differential equation with preconditioning $A^{-1}$,
\begin{equation}
\frac{\d \theta^t}{\d t}=-A(\theta^t)^{-1} F(\theta^t)
\end{equation}
initialized at $\theta^0=\theta_0$.

More precisely, for any $\eps>0$,
with probability $\geq 1-\eps$ one has
$\theta_{T/\deltat}-\theta^T=O(\sqrt{\deltat})$ when the constant in
$O()$ depends on $\eps$, $T$, $\lambda$, $\gamma$, $\sigma^2$, $R^2$, and the
derivatives of $F(\theta)$ and $A(\theta)$. The bounds are uniform for $T$ in
compact intervals.
\end{thm}

% Note: the variance assumption on $\hat A$ implies that the largest
% eivengalue of $A$ is at most $R^2$. In particular, $\gamma A$ has
% eigenvalues $\leq 1$.
The variance assumption on $\hat A$ directly controls the maximum
possible value via $\gamma\leq 1/R^2$, and, consequently,
the speed of convergence to
$A^{-1}$.
This assumption appears in
\cite{bachmoulines2013,defossezbach2015,dieuleveut2016} for $\hat
A=\tilde g\transp{\tilde g}$, where the value of $R^2$ for typical cases
is discussed.

With $\hat A=\tilde g\transp{\tilde g}$, 
the variance assumption on $\hat A$ is always satisfied with
$R^2=\sup \norm{\tilde g}^2$ if $\tilde g$ is bounded. \footnote{
TANGO uses $\hat A=(1-\deltat)\tilde g\transp{\tilde
g}+\frac{\deltat}{\gamma}\Id$ rather than $\hat A=\tilde g\transp{\tilde
g}$. Actually it is enough to check the assumption with $\tilde
g\transp{\tilde
g}$. Indeed one checks that if $\tilde g\transp{\tilde
g}$ satisfies the assumption with some $R^2$, then $(1-\deltat)\tilde
g\transp{\tilde
g}+\frac{\deltat}{\gamma}\Id$ satisfies the assumption with
$\max(R^2,1/\gamma)$, and that $\gamma\leq 1/R^2$ implies
$\gamma\leq 1/\max(R^2,1/\gamma)$.
} It is also
satisfied with
$R^2=\E\norm{\tilde g}^4/\lambda$, without bounded gradients. (Indeed, first, one has $\E \hat
A^2=\E (\norm{\tilde g}^2 \tilde g\transp{\tilde g})\leq (\sup
\norm{\tilde g}^2)\E \tilde g\transp{\tilde g}$; second, for any vector $u$, one has
$\transp{u}\E[\tilde g\transp{\tilde g}\tilde g\transp{\tilde
g}]u=\E[\transp{u}\tilde g\transp{\tilde g}\tilde g\transp{\tilde
g}u]\leq \E[\norm{u}^2\norm{\tilde g}^4]=\norm{u}^2\E\norm{\tilde g}^4$ while
$\transp{u}Au$ is at least $\lambda \norm{u}^2$.) If the distribution of
$\tilde g$ has bounded curtosis $\kappa$ in every direction, then the
assumption is satisfied with $R^2=\kappa \E\norm{\tilde g}^2$
\cite{dieuleveut2016}; in
particular,
for Gaussian $\tilde g$, with any covariance matrix, the assumption is
satisfied with $R^2=3\E\norm{\tilde g}^2$. All
these quantities can be estimated based on past values of $\tilde g$.

Theorem~\ref{thm:general} would still be valid with additional centered
noise on
$\theta$ and additional $o(\deltat)$ terms on $\theta$; for simplicity we
did not include them, as they are not needed for TANGO.

\begin{lem}
\label{lem:matrices}
Under assumptions of Theorem~\ref{thm:general}, the largest eigenvalue of
$A(\theta)$ is at most $R^2$. The operator
$(\Id-\gamma
A(\theta))$ is
$(1-\gamma\lambda)$-contracting.

Moreover, $\theta\mapsto A^{-1}(\theta)$ exists, is
bounded, and is $C^1$ with bounded derivatives. The same holds for $\theta\mapsto
A^{-1}(\theta)F(\theta)$.
\end{lem}

\begin{dem}
First, for any vector $u$, one has
$\norm{Au}^2=\norm{\E\hat Au}^2\leq \E \norm{\hat
Au}^2=\E[\transp{u}\transp{\hat A}\hat A u]\leq R^2
\transp{u}\!Au$. Taking $u$ an eigenvector associated with the largest
eigenvalue $\lambdamax$ of $A$ shows that $\lambdamax\leq R^2$.
Next, the eigenvalues of $A$ lie
between $\lambda$ and $R^2$ so that the eigenvalues of $\gamma A$ lie
between $\gamma\lambda$ and $1$. So the eigenvalues of $\Id-\gamma A$ lie
between $0$ and $1-\gamma\lambda$.%TODO: non-symmetric case?

Since $A$ is symmetric and its smallest eigenvalue is $\lambda>0$, it is
invertible with its inverse bounded by $1/\lambda$. Thanks to
$\partial_\theta A^{-1}=-A^{-1}(\partial_\theta A)A^{-1}$, the
derivatives of $A^{-1}$ are bounded.
\end{dem}

\begin{lem}
\label{lem:vnorm}
Under the notation and assumptions of Theorem~\ref{thm:general},
for any $k$,
\begin{equation}
\E \norm{v_k}^2\leq \frac{4\sigma^2}{\lambda^2}
\end{equation}
\end{lem}

Up to the factor $4$, this is optimal: indeed, when $\hat F$ and $\hat A$
have a distribution independent of $k$, the fixed point of $v$ in
expectation is $v=A^{-1}\E \hat F$, whose square norm is $\transp{(\E
\hat F)}A^{-2}\E \hat F$ which is $\norm{\E \hat F}^2/\lambda^2$ if $\E
\hat F$ lies in the
direction of the eigenvalue $\lambda$.

\begin{dem}
The proof is a variant of arguments appearing in \cite{bachmoulines2013};
in our case $A$ is not constant, $\hat F_k$ is not centered, $\hat
A_k$ is not rank-one, and we do
not use the norm associated with $A$ on the left-hand-side.
Let 
\begin{equation}
w_k\deq (\Id-\gamma\hat A_k)v_{k-1}
\end{equation}
so that $v_k=w_k+\gamma \hat F_k$. Consequently
\begin{equation}
\norm{v_k}^2=\norm{w_k}^2+\norm{\gamma \hat F_k}^2+2 \gamma w_k\cdot \hat
F_k \leq (1+\alpha)\norm{w_k}^2 + (1+1/\alpha)\norm{\gamma \hat F_k}^2
\end{equation}
for any $\alpha>0$,
thanks to $2ab=2(\sqrt{\alpha}\,a)(b/\sqrt{\alpha})\leq \alpha
a^2+b^2/\alpha$ for any $\alpha>0$ and $a,b\in \R$.

Now
\begin{equation}
\norm{w_k}^2=
\norm{v_{k-1}}^2-\gamma\transp{v_{k-1}}(\hat A_k +\transp{\hat A_k})v_{k-1}+
\gamma^2 \transp{v_{k-1}}\transp{\hat
A_k}\hat A_k v_{k-1}
\end{equation}

Take expectations conditionally to $\filtr_{k-1}$. Using $\E_{k-1}\left[
\transp{\hat A_k}\hat A_k\right] \preccurlyeq R^2 A_k$ we find
\begin{equation}
\E_{k-1}\norm{w_k}^2\leq
\norm{v_{k-1}}^2-\gamma(2-\gamma R^2)\transp{v_{k-1}}A_k v_{k-1}
\end{equation}
By the assumptions, $\gamma R^2\leq 1$ and $\transp{v_{k-1}}A_k
v_{k-1}\geq \lambda \norm{v_{k-1}}^2$. Thus
\begin{equation}
\E_{k-1}\norm{w_k}^2\leq
(1-\gamma\lambda)\norm{v_{k-1}}^2
\end{equation}
Taking $1+\alpha=\frac{1-\gamma\lambda/2}{1-\gamma\lambda}$ we find
\begin{align}
\E_{k-1}\norm{v_k}^2 &\leq
(1-\gamma\lambda/2)\norm{v_{k-1}}^2+(1+1/\alpha)\gamma^2\sigma^2
\\&\leq
(1-\gamma\lambda/2)\norm{v_{k-1}}^2 +
\frac{1-\gamma\lambda/2}{\gamma\lambda/2}\gamma^2\sigma^2
\end{align}
Taking unconditional expectations, we obtain
\begin{equation}
\E \norm{v_k}^2\leq (1-\gamma\lambda/2)\E\norm{v_{k-1}}^2+
\frac{1-\gamma\lambda/2}{\gamma\lambda/2}\gamma^2\sigma^2
\end{equation}
and by induction, starting at $v_0=0$, this implies
\begin{equation}
\E\norm{v_k}^2\leq
\frac{1-\gamma\lambda/2}{(\gamma\lambda/2)^2}\gamma^2\sigma^2\leq
\frac{4\sigma^2}{\lambda^2}
\end{equation}
\end{dem}

\begin{cor}
\label{cor:vbound}
Under the notation and assumptions of Theorem~\ref{thm:general}, for any
$n$, for any $\eps>0$, with probability $\geq 1-\eps$ one has
\begin{equation}
\sup_{0\leq k\leq n} \norm{v_k} \leq
\frac{2\sigma}{\lambda}\sqrt{\frac{n}\eps}
\end{equation}
\end{cor}

\begin{dem}
This follows from Lemma~\ref{lem:vnorm} by the Markov inequality and a
union bound.
\end{dem}

The next two lemmas result from standard martingale arguments; the
detailed proofs are given in the Appendix.

\begin{lem}
\label{lem:xibound}
Under the notation and assumptions of Theorem~\ref{thm:general},
let $\xi$ be the noise on $F$,
\begin{equation}
\xi_k\deq \hat F_k-F_k
\end{equation}

Let $(M_k)$ be any sequence of operators such that $M_k$ is
$\filtr_{k-1}$-measurable and $\opnorm{M_k}\leq \Lambda$ almost surely.

Then
\begin{equation}
\E \sum_{j=1}^n \norm{M_j\xi_j}^2\leq n\Lambda^2\sigma^2
\end{equation}
and moreover
for any
$n$, for any $\eps>0$, with probability $\geq 1-\eps$, for any $k\leq n$ one has
\begin{equation}
\norm{\sum_{j=k}^n M_j \xi_j}\leq 2\sqrt{\frac{n\Lambda^2\sigma^2}{\eps}}
\end{equation}
\end{lem}

\begin{lem}
\label{lem:zetabound}
Under the notation and assumptions of Theorem~\ref{thm:general}, 
set 
\begin{equation}
\zeta_k\deq (\hat A_k-A_k)v_{k-1}
\end{equation}

Let $(M_k)$ be any sequence of operators such that $M_k$ is
$\filtr_{k-1}$-measurable and $\opnorm{M_k}\leq \Lambda$ almost surely.
Let $\lambdamax=\sup_\theta \max \mathrm{eigenvalues}(A_k)$, which is
finite by Lemma~\ref{lem:matrices}.

Then
\begin{equation}
\E\sum_{j=1}^n \norm{M_j\zeta_j}^2\leq
4nR^2\lambdamax\Lambda^2\sigma^2/\lambda^2
\end{equation}
and moreover, for any
$n$, for any $\eps>0$, with probability $\geq 1-\eps$, for any $k\leq n$,
\begin{equation}
\norm{\sum_{j=k}^n M_j \zeta_j}\leq
4\sqrt{\frac{nR^2\lambdamax\Lambda^2\sigma^2}{\epsilon\lambda^2}}
\end{equation}
\end{lem}

\begin{dem}[ of Theorem~\ref{thm:general}]
Let $n\deq T/\deltat$ be the number of discrete steps corresponding to
continuous time $T$. All the constants implied in $O()$ notation below
depend on $T$ and on the assumptions of the theorem ($R^2$, $\gamma$,
$\lambda$, etc.), and we study the dependency on $\deltat$.

Similarly to Proposition~\ref{prop:nonstoch}, set
\begin{equation}
z_k\deq \theta_{k-1}-\deltat \gamma^{-1}B_k\,v_k
\end{equation}
where $B_k$ is a matrix to be defined later (equal to $A^{-1}$ for the
case of constant $A$).
Informally, $z$ contains $\theta$ plus the future updates to be
made to $\theta$ based on the current value of $v$.

Substituting
$\theta_{k-1}=\theta_{k-2}-\deltat \,v_{k-1}$
and
$v_k=v_{k-1} +
\gamma \hat F_k - \gamma A_kv_{k-1} - \gamma
\zeta_k$
into the definition of $z_k$, one finds
\begin{align}
z_k&=\theta_{k-2}-\deltat \, v_{k-1} -\deltat \gamma^{-1}
B_k\left(
v_{k-1} +
\gamma \hat F_k - \gamma A_kv_{k-1} - \gamma
\zeta_k\right)
\\&=\theta_{k-2}-\deltat B_k(\hat F_k-\zeta_k)
-\deltat\left(\Id+\gamma^{-1}B_k-B_kA_k\right)v_{k-1}
\\&=z_{k-1}-\deltat B_k(\hat F_k-\zeta_k)
-\deltat\left(\Id-B_kA_k+\gamma^{-1}(B_k-B_{k-1})\right)v_{k-1}
\label{eq:zupdate}
\end{align}

Now define $B_k$ in order to cancel the $v_{k-1}$ term, namely
\begin{equation}
B_{k-1}\deq B_k+\gamma(\Id-B_kA_k)
\end{equation}
initialized with $B_n\deq A_n^{-1}$. (If $A$ is constant, then
$B=A^{-1}$.) Then $\deltat \gamma^{-1} B_kv_k$ represents all the future updates to
$\theta$ stemming from the current value $v_k$.

With this choice, the update for $z$ is
\begin{equation}
z_k=z_{k-1}-\deltat B_k(\hat F_k-\zeta_k)=z_{k-1}-\deltat
B_k(F_k+\xi_k-\zeta_k)
\end{equation}
Remove the noise by defining
\begin{equation}
y_k\deq z_k-\deltat \sum_{j=k+1}^n B_j(\xi_j-\zeta_j)
\end{equation}
so that
\begin{equation}
y_k=y_{k-1}-\deltat B_k F_k
\end{equation}

Assume for now that $B_k=A^{-1}(\theta_{k-1})+O(\sqrt{\deltat})$.
Then
\begin{equation}
y_k
=y_{k-1}-\deltat
A^{-1}(\theta_{k-1})F(\theta_{k-1})+O(\deltat^{3/2})
\end{equation}

Since $A^{-1}F$ is Lipschitz (Lemma~\ref{lem:matrices}), we have
\begin{equation}
y_k=y_{k-1}-\deltat A^{-1}(y_{k-1})F(y_{k-1})+O(\deltat
\norm{y_{k-1}-\theta_{k-1}})+O(\deltat^{3/2})
\end{equation}

If we prove that $y_{k-1}-\theta_{k-1}=O(\sqrt{\deltat})$ then we find
\begin{equation}
y_k=y_{k-1}-\deltat A^{-1}(y_{k-1})F(y_{k-1})+O(\deltat^{3/2})
\end{equation}
so that $y_k$ is a Euler numerical scheme for the differential equation
$\dot y=-A^{-1}(y)F(y)$, and thus converges to the natural gradient
trajectory up to $O(\sqrt{\deltat})$, uniformly on the time
interval $[0;T]$.

Since we assumed that $\theta_k-y_k=O(\sqrt{\deltat})$, this holds for
$\theta_k$ as well.
% Since
% $\theta_k-y_k=\theta_k-z_k+z_k-y_k=O(\deltat
% \opnorm{B_k}\norm{v_k})+O(\sqrt{\deltat})=O(\deltat\sqrt{n})+O(\sqrt{\deltat})=O(\sqrt{\deltat})$ by
% Corollary~\ref{cor:vbound} (and because $B_k=A_k^{-1}+O(\sqrt{\deltat})$ is
% bounded), $\theta_k$ converges to the natural gradient trajectory as
% well.

We still have to prove the two assumptions that $y_{k-1}-\theta_{k-1}=O(\sqrt{\deltat})$ and
that $B_k=A^{-1}(\theta_{k-1})+O(\sqrt{\deltat})$.

\begin{lem}
\label{lem:Bmatrix}
Define $B_{k-1}\deq B_k+\gamma(\Id-B_kA_k)$ initialized with $B_n\deq
A_n^{-1}$. Then for any $\eps>0$, with probability $\geq 1-\eps$, one has
$\sup_k \opnorm{B_k-A_k^{-1}}=O(\sqrt{\deltat})$.
\end{lem}

\begin{dem}[ of Lemma~\ref{lem:Bmatrix}]
With this definition one has
\begin{equation}
B_{k-1}-A^{-1}_{k-1}=(B_k-A_k^{-1})(\Id-\gamma A_k)+A_k^{-1}-A_{k-1}^{-1}
\end{equation}
by a direct computation.

Now
$A_k^{-1}-A^{-1}_{k-1}=A^{-1}(\theta_{k-1})-A^{-1}(\theta_{k-2})=O(\theta_{k-1}-\theta_{k-2})$
because $A^{-1}$ is Lipschitz. Moreover
$\theta_{k-1}=\theta_{k-2}-\deltat v_{k-1}$. So
$A_k^{-1}-A^{-1}_{k-1}=O(\deltat \norm{v_{k-1}})$. Thanks to
Corollary~\ref{cor:vbound}, with probability $\geq
1-\eps$, $\sup_k \norm{v_{k-1}}=O(\sqrt{n})=O(1/\sqrt{\deltat})$ so that
$A_k^{-1}-A^{-1}_{k-1}$ is $O(\sqrt{\deltat})$, uniformly in $k$.

Now, the operator $(\Id-\gamma A_k)$ is
$(1-\gamma\lambda)$-contracting. Therefore,
\begin{equation}
\opnorm{B_{k-1}-A^{-1}_{k-1}}\leq
(1-\gamma\lambda)\opnorm{B_k-A_k^{-1}}+O(\sqrt{\deltat})
\end{equation}
and $B_n-A_n^{-1}$ is $0$, so by induction,
$\opnorm{B_{k-1}-A^{-1}_{k-1}}=O(\sqrt{\deltat})$, uniformly in $k$.
\end{dem}

Back to the proof of Theorem~\ref{thm:general}.
To prove that $y_k-\theta_k=O(\sqrt{\deltat})$, let us first prove that
$y_k-z_k=O(\sqrt{\deltat})$. We have
\begin{equation}
z_k-y_k=\deltat\sum_{j=k+1}^n B_j(\xi_j-\zeta_j)
\end{equation}
Thanks to Lemma~\ref{lem:Bmatrix}, this rewrites as
\begin{equation}
z_k-y_k=\deltat\sum_{j=k+1}^n A^{-1}_j
(\xi_j-\zeta_j)+O\left(\deltat^{3/2} \sum_{j=k+1}^n
(\norm{\xi_j}+\norm{\zeta_j})\right)
\end{equation}
For the first term, note that $A^{-1}_j=A^{-1}(\theta_{j-1})$ is
$\filtr_{j-1}$-measurable (while $B_j$ is not, because it depends on
$\theta_k$ for $k\geq j$). By
Lemmas~\ref{lem:xibound} and \ref{lem:zetabound},
$\sum A_j \xi_j$ and $\sum
A_j \zeta_j$ are both $O(\sqrt{n})=O(\sqrt{1/\deltat})$ with high
probability. So the first term of $z_k-y_k$ is $O(\sqrt{\deltat})$.

For the second term,
% the martingale argument does not apply, as the
% $O(\sqrt{\deltat})$ term is correlated to $\xi_j$ and $\zeta_j$ (because
% $B_j$ depends on all $\theta_k$'s for $k\geq j$). But
\begin{align}
% \norm{\sum_{j=k}^n O(\sqrt{\deltat})\xi_j}&\leq 
% \sum_{j=k}^n O(\sqrt{\deltat})\norm{\xi_j}=
% O\left(\sqrt{\deltat}\right)\sum_{j=k}^n \norm{\xi_j}
% \\&\leq O(\sqrt{\deltat})\sum_{j=1}^n \norm{\xi_j}
% \leq O(\sqrt{\deltat})\sqrt{n}\sqrt{\sum_{j=1}^n \norm{\xi_j}^2}
\sum_{j=k+1}^n \norm{\xi_j}
\leq \sum_{j=1}^n \norm{\xi_j}
\leq \sqrt{n}\sqrt{\sum_{j=1}^n \norm{\xi_j}^2}
\end{align}
by Cauchy--Schwarz. By Lemma~\ref{lem:xibound}, 
$\E\sum \norm{\xi_j}^2$ is $O(n)$. So with probability $\geq
1-\eps$, thanks to the Markov inequality, $\sqrt{\sum \norm{\xi_j}^2}$ is
$O(\sqrt{n})$ where the constant in $O()$ depends on $\eps$.
Therefore, $\sum_{j=k}^n \norm{\xi_j}$ is
$O(n)=O(1/\deltat)$.  
The
same argument applies to $\zeta$ thanks to Lemma~\ref{lem:zetabound}.

Therefore, 
$z_k-y_k$ is
$O(\sqrt{\deltat})$.

Finally, $z_k-\theta_k$ is $O(\deltat \norm{v_k})$ which is
$O(\deltat\sqrt{n})=O(\sqrt{\deltat})$ by Corollary~\ref{cor:vbound}.
Therefore $y_k-\theta_k$ is $O(\sqrt{\deltat})$ as well.

\end{dem}

\appendix
\section{Additional proofs}
%\section{Proof of Proposition~\ref{prop:trajav}}
\label{sec:trajavproof}

\begin{dem}[ of Proposition~\ref{prop:trajav}]
Start with the algorithm in Proposition~\ref{prop:trajav}, with any noise
$\xi_k$.
Under the update for $\theta_k$ one has
\begin{equation}
\label{eq:thetadiff}
\theta_k-\thetaf_k=(1-\deltat_k)(\theta_{k-1}-\thetaf_k)
\end{equation}
Now set
\begin{equation}
v_k\deq \theta_{k-1}-\thetaf_k
\end{equation}
so that the update for $\theta_k$ is
$\theta_k=\theta_{k-1}-\deltat_k\theta_{k-1}+\deltat_k \thetaf_k=\theta_{k-1}-\deltat_k
v_k$ by construction. To determine the update for $v$, remove
$\theta_{k-1}$
from the update of $\thetaf_k$:
\begin{equation}
\label{eq:partialvupdate}
\thetaf_{k}-\theta_{k-1}=\thetaf_{k-1}-\theta_{k-1}-\gamma \gf_k+\gamma\xi_k
\end{equation}
where we abbreviate $\gf_k\deq \frac{\partial \ell(y_k|x_k)}{\partial
\thetaf_{k-1}}$, the gradient of the loss at $\thetaf_{k-1}$.

Let $H_k$ be the Hessian of the loss on the $k$-th example with respect
to the parameter. Since losses
are quadratic, the gradient of the loss is a linear function of the
parameter:
\begin{equation}
\gf_k=g_k+H_k (\thetaf_{k-1}-\theta_{k-1})
\end{equation}
where $g_k\deq \frac{\partial \ell(y_k|x_k)}{\partial
\theta_{k-1}}$ is the gradient of the loss at $\theta_{k-1}$.

Thus \eqref{eq:partialvupdate} rewrites as
\begin{equation}
v_k=-\thetaf_{k-1}+\theta_{k-1}+\gamma g_k +\gamma
H_k(\thetaf_{k-1}-\theta_{k-1})-\gamma\xi_k
\end{equation}
and thanks to \eqref{eq:thetadiff},
\begin{equation}
\theta_{k-1}-\thetaf_{k-1}=(1-\deltat_{k-1})v_{k-1}
\end{equation}
so the above rewrites as
\begin{equation}
v_k=(1-\deltat_{k-1})v_{k-1}+\gamma g_k-\gamma (1-\deltat_{k-1})H_k
v_{k-1}-\gamma\xi_k
\end{equation}

If we set
\begin{equation}
\xi_k\deq (1-\deltat_{k-1})(\tilde g_k\transp{\tilde g_k}-H_k)v_{k-1}
\end{equation}
then this is identical to TANGO. However, we still have to prove that
such a $\xi_k$
is a centered noise, namely, $\E\xi_k=0$. This will be the case if
\begin{equation}
H_k=\E\tilde g_k\transp{\tilde g_k}
\end{equation}
where the expectation is with respect to the choice of the random output
$\tilde y_k$ given $x_k$. From the double definition of the Fisher
matrix of a probabilistic model, we know that
\begin{equation}
\E_{\tilde y\sim p_\theta(\tilde y|x)} \frac{\partial \ell(\tilde y|x)}{\partial \theta}
\transp{\frac{\partial \ell(\tilde y|x)}{\partial \theta}}=\E_{\tilde y\sim
p_\theta(\tilde y|x)}\frac{\partial^2\ell(\tilde y|x)}{\partial \theta^2}
\end{equation}
Since we have assumed that this Hessian does not depend on $\tilde y$, it is
equal to $H_k$.

Thus TANGO rewrites as averaged SGD with a particular model of
noise on the fast parameter.
\end{dem}

\begin{dem}[ of Lemma~\ref{lem:xibound}]
This is a standard martingale argument.
By the variance assumption on
$\hat F_k$, one has $\E_{k-1}\norm{\xi_k}^2\leq
\sigma^2$. Likewise, $\E_{k-1}\norm{M_k \xi_k}^2 \leq \Lambda^2\sigma^2$.
This proves the first claim.

Moreover, since $\E_{k-1}\xi_k=0$ and $M_k$ is $\filtr_{k-1}$-measurable,
$\E_{k-1} M_k\xi_k=0$, namely,
the $M_k \xi_k$
are martingale increments.

Setting 
$X_k\deq \norm{\sum_{j=1}^k M_j \xi_j}^2$, we find
$\E_k X_{k+1}=X_k+2\E_k [(M_{k+1}\xi_{k+1})\cdot
\sum_{j=1}^k M_j \xi_j ]+\E_k \norm{M_{k+1} \xi_{k+1}}^2=X_k+\E_k
\norm{M_{k+1} \xi_{k+1}}^2$.

Consequently, $\E X_n\leq
n\Lambda^2\sigma^2$.
Moreover, $\E_k X_{k+1}\geq
X_k$, so that $X_k$ is a submartingale. Therefore, by Doob's martingale inequality, with probability $\geq
1-\eps$,
\begin{equation}
\sup_{0\leq k\leq n} X_k \leq \frac{\E X_n}{\eps}\leq
\frac{n\Lambda^2\sigma^2}{\eps}
\end{equation}

Finally, $\sum_{j=k}^n M_j\xi_j =\sum_{j=1}^n
M_j\xi_j-\sum_{j=1}^{k-1}M_j\xi_j$,
hence the conclusion by the triangle inequality.
\end{dem}

\begin{dem}[ of Lemma~\ref{lem:zetabound}]
The argument is similar to the preceding lemma, together with the bound
on $\E\norm{v_k}^2$ from Lemma~\ref{lem:vnorm}. Conditionally to
$\filtr_{k-1}$ one has
$\E_{k-1} \norm{\zeta_k}^2=\E_{k-1} \transp{v_{k-1}}(\hat A_k -A_k)(\hat
A_k-A_k)v_{k-1}
=\E_{k-1}\transp{v_{k-1}}\hat A_k^2v_{k-1}- \transp{v_{k-1}}A_k^2v_{k-1}
\leq R^2 \transp{v_{k-1}}A_k v_{k-1}
\leq R^2\lambdamax \norm{v_{k-1}}^2
$. Therefore, $\E \norm{\zeta_k}^2\leq
R^2\lambdamax\E\norm{v_{k-1}}^2\leq 4R^2\lambdamax \sigma^2/\lambda^2$ by
Lemma~\ref{lem:vnorm}.

The operators $M_k$ introduce an additional
factor $\Lambda^2$. Consequently, $\E \sum_{k=1}^n
\norm{M_k\zeta_k}^2\leq 4nR^2\Lambda^2\lambdamax\sigma^2/\lambda^2$.

The rest of the proof is identical to
Lemma~\ref{lem:xibound}.
% 
% Setting $Z_k \deq \norm{\sum_{j=1}^k\zeta_k}^2$ and using
% $\E[\zeta_{k+1}\mid \filtr_k]=0$, one has, as before,
% $\E[Z_{k+1}\mid \filtr_k]=Z_k+\E[\norm{\zeta_{k+1}}^2\mid \filtr_k]$
% which lies between $Z_k$ and $Z_k+R^2\lambdamax \norm{v_k}^2$.
% Therefore, $Z_k$ is a supermartingale. Besides, by induction, $\E Z_k\leq
% R^2\lambdamax\sum_{j=1}^{k-1}\E\norm{v_k}^2$.
% 
% Thus by Lemma~\ref{lem:vnorm},
% $\E Z_n \leq \frac{4nR^2\lambdamax\sigma^2}{\lambda^2}
% $.
% 
% The conclusion follows by Doob's martingale inequality, as before.
\end{dem}

\bibliographystyle{alpha}
\bibliography{tango}

\newcommand{\etalchar}[1]{$^{#1}$}
\begin{thebibliography}{LBOM98}

\bibitem[ABH16]{agarwal2016secondorderlineartime}
Naman Agarwal, Brian Bullins, and Elad Hazan.
\newblock Second order stochastic optimization in linear time.
\newblock {\em arXiv preprint arXiv:1602.03943}, 2016.

\bibitem[Ama98]{Amari1998}
Shun-ichi Amari.
\newblock Natural gradient works efficiently in learning.
\newblock {\em Neural Comput.}, 10:251--276, February 1998.

\bibitem[BM13]{bachmoulines2013}
Francis Bach and Eric Moulines.
\newblock Non-strongly-convex smooth stochastic approximation with convergence
  rate o (1/n).
\newblock In {\em Advances in neural information processing systems}, pages
  773--781, 2013.

\bibitem[DB15]{defossezbach2015}
Alexandre D{\'e}fossez and Francis Bach.
\newblock Averaged least-mean-squares: Bias-variance trade-offs and optimal
  sampling distributions.
\newblock In {\em Artificial Intelligence and Statistics}, pages 205--213,
  2015.

\bibitem[DFB16]{dieuleveut2016}
Aymeric Dieuleveut, Nicolas Flammarion, and Francis Bach.
\newblock Harder, better, faster, stronger convergence rates for least-squares
  regression.
\newblock {\em arXiv preprint arXiv:1602.05419}, 2016.

\bibitem[DPCB13]{desjardins2013metricfreeNG}
Guillaume Desjardins, Razvan Pascanu, Aaron Courville, and Yoshua Bengio.
\newblock Metric-free natural gradient for joint-training of boltzmann
  machines.
\newblock {\em arXiv preprint arXiv:1301.3545}, 2013.

\bibitem[DSP{\etalchar{+}}15]{desjardins2015NNN}
Guillaume Desjardins, Karen Simonyan, Razvan Pascanu, et~al.
\newblock Natural neural networks.
\newblock In {\em Advances in Neural Information Processing Systems}, pages
  2071--2079, 2015.

\bibitem[KB17]{karmakar2017twotimescaletd}
Prasenjit Karmakar and Shalabh Bhatnagar.
\newblock Two time-scale stochastic approximation with controlled markov noise
  and off-policy temporal-difference learning.
\newblock {\em Mathematics of Operations Research}, 2017.

\bibitem[KP92]{KloedenPlaten92}
Peter~E. Kloeden and Eckhard Platen.
\newblock {\em Numerical solution of stochastic differential equations},
  volume~23 of {\em Applications of Mathematics (New York)}.
\newblock Springer-Verlag, Berlin, 1992.

\bibitem[LBOM98]{lecun98efficientbackprop}
Yann {Le Cun}, L\'{e}on Bottou, Genevieve~B. Orr, and Klaus-Robert
  M{\"{u}}ller.
\newblock Efficient backprop.
\newblock In {\em Neural Networks, Tricks of the Trade}, Lecture Notes in
  Computer Science LNCS~1524. Springer Verlag, 1998.

\bibitem[LMB07]{TONGA}
Nicolas {Le Roux}, Pierre{-}Antoine Manzagol, and Yoshua Bengio.
\newblock Topmoumoute online natural gradient algorithm.
\newblock In {\em Advances in Neural Information Processing Systems 20,
  Proceedings of the Twenty-First Annual Conference on Neural Information
  Processing Systems, Vancouver, British Columbia, Canada, December 3-6, 2007},
  pages 849--856, 2007.

\bibitem[Mar10]{Martens2010}
James Martens.
\newblock Deep learning via {H}essian-free optimization.
\newblock In Johannes F{\"{u}}rnkranz and Thorsten Joachims, editors, {\em
  Proceedings of the 27th International Conference on Machine Learning
  (ICML-10), June 21-24, 2010, Haifa, Israel}, pages 735--742. Omnipress, 2010.

\bibitem[Mar14]{martens2014natgradinsights}
James Martens.
\newblock New insights and perspectives on the natural gradient method.
\newblock {\em arXiv preprint arXiv:1412.1193}, 2014.

\bibitem[MB11]{bachmoulines2011}
\'{E}ric Moulines and Francis~R Bach.
\newblock Non-asymptotic analysis of stochastic approximation algorithms for
  machine learning.
\newblock In {\em Advances in Neural Information Processing Systems}, pages
  451--459, 2011.

\bibitem[MCO16]{marceauollivier2016}
Ga{\'e}tan Marceau-Caron and Yann Ollivier.
\newblock Practical riemannian neural networks.
\newblock {\em arXiv preprint arXiv:1602.08007}, 2016.

\bibitem[MG15]{martensgrosse2015}
James Martens and Roger Grosse.
\newblock Optimizing neural networks with kronecker-factored approximate
  curvature.
\newblock In {\em International Conference on Machine Learning}, pages
  2408--2417, 2015.

\bibitem[MS11]{Martens2011}
James Martens and Ilya Sutskever.
\newblock Learning recurrent neural networks with {H}essian-free optimization.
\newblock In {\em ICML}, pages 1033--1040, 2011.

\bibitem[MS12]{Martens2012}
James Martens and Ilya Sutskever.
\newblock Training deep and recurrent neural networks with {H}essian-free
  optimization.
\newblock In Grégoire Montavon, Geneviève~B. Orr, and Klaus-Robert Müller,
  editors, {\em Neural Networks: Tricks of the Trade}, volume 7700 of {\em
  Lecture Notes in Computer Science}, pages 479--535. Springer, 2012.

\bibitem[Oll15]{Ollivier2015}
Yann Ollivier.
\newblock Riemannian metrics for neural networks {I}: feedforward networks.
\newblock {\em Information and Inference}, 4(2):108--153, 2015.

\bibitem[Oll17]{natkal}
Yann Ollivier.
\newblock Online natural gradient as a kalman filter.
\newblock {\em arXiv preprint arXiv:1703.00209}, 2017.

\bibitem[PJ92]{polyakjuditsky1992}
Boris~T Polyak and Anatoli~B Juditsky.
\newblock Acceleration of stochastic approximation by averaging.
\newblock {\em SIAM Journal on Control and Optimization}, 30(4):838--855, 1992.

\bibitem[Rup88]{ruppert1988}
David Ruppert.
\newblock Efficient estimations from a slowly convergent robbins-monro process.
\newblock Technical report, Cornell University Operations Research and
  Industrial Engineering, 1988.

\bibitem[Tad04]{tadic2004twoscales}
Vladislav~B Tadic.
\newblock Almost sure convergence of two time-scale stochastic approximation
  algorithms.
\newblock In {\em American Control Conference, 2004. Proceedings of the 2004},
  volume~4, pages 3802--3807. IEEE, 2004.

\end{thebibliography}

\end{document}